\documentclass[letterpaper]{article}
\usepackage{uai2018}
\usepackage[margin=1in]{geometry}

\usepackage{times}
\usepackage{amsmath}
\usepackage{amsfonts}
\usepackage{graphicx}
\usepackage{subfig}
\usepackage{float}

\title{Per-decision Multi-step Temporal Difference Learning with Control Variates}

\author{
	{\bf Kristopher De Asis} \\
	Department of Computing Science \\
	University of Alberta \\
	Edmonton, AB T6G 2E8 \\
	\textit{kldeasis@ualberta.ca} \\
	\And
	{\bf Richard S. Sutton} \\
	Department of Computing Science \\
	University of Alberta \\
	Edmonton, AB T6G 2E8 \\
	\textit{rsutton@ualberta.ca}
}

\begin{document}

\maketitle

\begin{abstract}
Multi-step temporal difference (TD) learning is an important approach in reinforcement learning, as it unifies one-step TD learning with Monte Carlo methods in a way where intermediate algorithms can outperform either extreme. They address a bias-variance trade off between reliance on current estimates, which could be poor, and incorporating longer sampled reward sequences into the updates. Especially in the off-policy setting, where the agent aims to learn about a policy different from the one generating its behaviour, the variance in the updates can cause learning to diverge as the number of sampled rewards used in the estimates increases. In this paper, we introduce per-decision control variates for multi-step TD algorithms, and compare them to existing methods. Our results show that including the control variates can greatly improve performance on both on and off-policy multi-step temporal difference learning tasks.
\end{abstract}

\section{TEMPORAL DIFFERENCE LEARNING}
\label{sec:intro}

\textit{Temporal-difference} (TD) methods (Sutton, 1988) combine ideas from Monte Carlo and dynamic programming methods, and are an important approach in \textit{reinforcement learning}. They allow learning to occur from raw experience in the absence of a model of the environment's dynamics, like with Monte Carlo methods, while computing estimates which bootstrap off of other estimates, like with dynamic programming. TD methods provide a way to learn online and incrementally in both prediction and control settings.

Several TD methods have been proposed. Sarsa (Rummery \& Niranjan, 1994; Sutton, 1996) is a classical \textit{on-policy} algorithm, where the policy being learned about, the \textit{target policy}, is identical to the one generating the behaviour, the \textit{behaviour policy}. However, Sarsa can be extended to learn \textit{off-policy}, where the target policy can differ from the behaviour policy, through the use of per-decision \textit{importance sampling} (Precup et al., 2000). Expected Sarsa (van Seijen et al., 2009) is another extension of Sarsa where instead of using the value of the current state-action pair to update the value of the previous state, it uses the expectation of the values of all actions in the current state under the target policy. Since Expected Sarsa takes the expectation under the target policy, it can be used off-policy without importance sampling to correct for the discrepancy between its target and behaviour policies. $Q$-learning  (Watkins, 1989) is arguably the most popular off-policy TD control algorithm, as it can also perform off-policy learning without importance sampling, but it is equivalent to Expected Sarsa where the target policy is greedy. The above methods are often described in the one-step case, but they can be extended across multiple time steps.

Multi-step TD methods, such as the $n$-step TD and TD($\lambda$) methods, create a spectrum of algorithms where at one end exists one-step TD learning, and at the other, exists Monte Carlo Methods. Intermediate algorithms are created which, due to a bias-variance tradeoff, can outperform either extreme (Jaakkola et al., 1994). Multi-step off-policy algorithms, especially ones with explicit use of importance sampling, have significantly larger variance than their on-policy counterparts (Sutton \& Barto, 1998), and several proposals have been made to address this issue in the TD($\lambda$) space of algorithms (Munos et al., 2016; Mahmood et al., 2017).

In this paper, we focus on $n$-step TD algorithms as they provide exact computation of the multi-step return, have conceptual clarity, and provide the foundation for TD($\lambda$) methods. We formulate per-decision control variates for existing $n$-step TD algorithms, and give insight on their implications in the TD($\lambda$) space of algorithms. On problems with tabular representations as well as one with function approximation, we show that the introduction of per-decision control variates can improve the performance of existing $n$-step TD methods on both on and off-policy prediction and control tasks.

\section{ONE-STEP TD METHODS}
\label{sec:onesteptd}
The sequential decision-making problem in reinforcement learning is often modeled as a \textit{Markov decision process} (MDP). Under the MDP framework, an \textit{agent} interacts with an environment over a sequence of discrete time steps. At each time step $t$, the agent receives information about the environment's current \textit{state}, $S_t \in \mathcal{S}$, where $\mathcal{S}$ is the set of all possible states in the MDP. The agent is to use this state information to select an \textit{action}, $A_t \in \mathcal{A}(S_t)$, where $\mathcal{A}(s)$ is the set of possible actions in state $s$. Based on the environment's current state and the agent's selected action, the agent receives a \textit{reward}, $R_{t+1} \in \mathbb{R}$, and gets information about the environment's next state, $S_{t+1} \in \mathcal{S}$, according to the \textit{environment model}:
$p(r,s'|s,a)={P(R_{t+1}=r,S_{t+1}=s'|S_t=s,A_t=a)}$.

The agent selects actions according to a \textit{policy}, $\pi(s,a) = P(A_t=a|S_t=s)$, which gives a probability distribution across actions $a \in \mathcal{A}(s)$ for a given state $s$. Through policy iteration (Sutton \& Barto, 1998), the agent can learn an optimal policy, $\pi^*$, where behaving under it will maximize the expected discounted return:
\begin{equation}
G_t = R_{t+1} + \gamma R_{t+2} + \gamma^2 R_{t+3} + ... = \sum_{k=0}^{T-t - 1} \gamma^{k} R_{t+k+1}
\end{equation}
given a discount factor $\gamma \in [0,1]$ and $T$ equal to the final time step in an episodic task, or $\gamma \in [0,1)$ and $T$ equal to infinity for a continuing task.

\textit{Value-based methods} approach the sequential decision-making problem by computing \textit{value functions}, which provide estimates of what the return will be from a particular state onwards. In prediction problems, also referred to as \textit{policy evaluation}, the goal is to estimate the return under a particular policy as accurately as possible, and a \textit{state-value function} is often estimated. It is defined to be the expected return when starting in state $s$ and following policy $\pi$: $v_\pi(s) = \mathbb{E}_\pi[G_t|S_t=s]$. For control problems, the policy which maximizes the expected return is to be learned, and an \textit{action-value function} from which a policy can be derived is instead estimated. It is defined to be the expected return when taking action $a$ in state $s$, and following policy $\pi$:
\begin{equation}
q_\pi(s,a) = \mathbb{E}_\pi[G_t|S_t=s,A_t=a]
\label{eqn:qdef}
\end{equation}
Of note, the action-value function can still be used for prediction problems, and the state-value can be computed as an expectation across action-values under the policy $\pi$ for a given state:
\begin{equation}
v_\pi(s) = \mathbb{E}_{\pi}[q_\pi(s,\cdot)] = \sum_{a}{\pi(s, a)q_\pi(s,a)}
\label{eqn:qtov}
\end{equation}
One-step TD methods learn an approximate value function, such as $Q \approx q_\pi$ for action-values, by computing an estimate of the return, $\hat{G}_t$. First, Equation \ref{eqn:qdef} can be written in terms of its succesor state-action pairs, also known as the \textit{Bellman equation} for $q_\pi$:
\begin{equation}
q_\pi(s,a) = \sum_{r,s'}{p(r,s'|s,a)\Big(r + \gamma\sum_{a'}\pi(s', a')q_\pi(s',a')\Big)}
\label{eqn:qbellman}
\end{equation}
Based on Equation \ref{eqn:qbellman}, one-step TD methods estimate the return by taking an action in the environment according to a policy, sampling the immediate reward, and bootstrapping off of the current estimates in the value function for the remainder of the return. The difference between this \textit{TD target} and the value of the previous state-action pair is then computed, and is often referred to as the \textit{TD error}. The previous state-action pair's value is then updated by taking a step proportional to the TD error with step size $\alpha \in (0, 1]$:
\begin{align}
\hat{G}_t &= R_{t+1} + \gamma Q(S_{t+1},A_{t+1})
\label{eqn:tdtarget_sarsa} \\
Q(S_t,A_t) &\leftarrow Q(S_t,A_t) + \alpha [\hat{G}_t - Q(S_t, A_t)]
\label{eqn:tdupdate}
\end{align}
Equations \ref{eqn:tdtarget_sarsa} and \ref{eqn:tdupdate} correspond to the Sarsa algorithm. It can be seen that in state $S_{t+1}$, it samples an action $A_{t+1}$ according to its behaviour policy, and then bootstraps off of the value of this state-action pair. With a sufficiently small step size, this estimates the expectation under its behaviour policy over the values of successor state-action pairs in Equation \ref{eqn:qbellman}, allowing for on-policy learning.

In the off-policy case, the discrepancy from $A_{t+1}$ being drawn from the behaviour policy needs to be corrected. One approach is to correct the affected terms with per-decision \textit{importance sampling}. With actions sampled from a behaviour policy $\mu$, and a target policy $\pi$, the estimate of the return of off-policy Sarsa with per-decision importance sampling becomes:
\begin{align}
\rho_t &= \frac{\pi(S_t, A_t)}{\mu(S_t, A_t)}
\label{eqn:rho} \\
\hat{G}_t &= R_{t+1} + \gamma \rho_{t+1}Q(S_{t+1},A_{t+1})
\label{eqn:tdtarget_offsarsa}
\end{align}
Note that in the on-policy case, $\rho_t$ is always 1, strictly generalizing the original on-policy TD target in Equation \ref{eqn:tdtarget_sarsa}.

Another approach for the off-policy case is to compute the expectation of all successor state action pairs under the target policy directly, instead of sampling and correcting the discrepancy. This approach has lower variance and is often preferred in the one-step setting for action-values, and gives the \textit{Expected Sarsa} algorithm (van Seijen et al., 2009) characterized by the following TD target:
\begin{align}
\hat{G}_t &= R_{t+1} + \gamma \mathbb{E}_{\pi}[Q(S_{t+1},\cdot)]
\label{eqn:tdtarget_expsarsa}
\end{align}

\section{MULTI-STEP TD LEARNING}
\label{sec:multisteplearning}
TD algorithms are referred to as one-step TD algorithms when they only incorporate information from a single time step in the estimate of the return that the value function is being updated towards. In multi-step TD methods, a longer sequence of experienced rewards is used to estimate the return. For example, on-policy $n$-step Sarsa would update an action-value $Q(S_t, A_t)$ towards the following estimate:
\begin{align}
\nonumber
\hat{G}_{t:t+n} &= R_{t+1} + \gamma R_{t+2} + ... + \gamma^n Q(S_{t+n}, A_{t+n}) \\
&= \sum_{k=0}^{n - 1}{\gamma^k R_{t + k + 1}} + \gamma^n Q(S_{t+n}, A_{t+n})
\label{eqn:nstepsarsa}
\end{align}
Of note, $n$-step Expected Sarsa (Sutton \& Barto, 2018) is identical up until the $n$-th step, where it instead bootstraps off of the expectation under the target policy:
\begin{equation}
\hat{G}_{t:t+n} = \sum_{k=0}^{n - 1}{\gamma^k R_{t + k + 1}} + \gamma^n \mathbb{E}_{\pi}[Q(S_{t+n}, \cdot)]
\label{eqn:nstepexpsarsa}
\end{equation}
The $n$-step returns can also be written recursively, and is convenient in the more general per-decision off-policy case. If we define the following bootstrapping condition:
\begin{equation}
\hat{G}_{t:t} = Q(S_t, A_t)
\label{eqn:recnbootstrap}
\end{equation}
The $n$-step extension of off-policy Sarsa with per-decision importance sampling, as characterized by Equations \ref{eqn:rho} and \ref{eqn:tdtarget_offsarsa}, can now be written as:
\begin{equation}
\hat{G}_{t:t+n} = R_{t+1} + \gamma \rho_{t+1} \hat{G}_{t+1:t+n}
\label{eqn:recnsarsa}
\end{equation}

TD algorithms which update towards these $n$-step estimates of the return constitute the $n$-step TD algorithm family (Sutton \& Barto, 2018). Their computational complexity increases with $n$, but have the benefit of conceptual clarity, and exact computation of the multi-step return. The $n$-step returns also provide the foundation for other multi-step TD algorithms.

Another family of multi-step per-decision TD algorithms, \textit{TD($\lambda$)}, is also used in practice. They are characterized by computing a geometrically weighted sum of $n$-step returns, denoted as the \textit{$\lambda$-return}:
\begin{equation}
\hat{G}_t^\lambda = (1 - \lambda)\sum_{n=1}^{\infty}{\lambda^{n - 1}\hat{G}_{t:t+n}}
\label{eqn:lambdareturn}
\end{equation}
It introduces a hyperparameter $\lambda \in [0, 1]$ where $\lambda = 0$ gives one-step TD, and increasing $\lambda$ effectively increases the number of sampled rewards included in the estimated return. Substituting the $n$-step Sarsa return (\ref{eqn:recnsarsa}) into Equation \ref{eqn:lambdareturn} gives the $\lambda$-return for the Sarsa($\lambda$) algorithm, and assuming $Q$ does not change, it can be expressed as a sum of one-step Sarsa's TD errors:
\begin{align}
\nonumber
\hat{G}_{t} &= R_{t+1} + \gamma \rho_{t+1} Q(S_{t+1}, A_{t+1}) \\
\hat{G}_t^\lambda &= Q(S_t, A_t) + \sum_{k=t}^{\infty}{(\hat{G}_k - Q(S_k, A_k))\prod_{i=t+1}^{k}{\gamma \lambda \rho_i}}
\label{eqn:sumlsarsa}
\end{align}
This shows that the $\lambda$-return for Sarsa($\lambda$) can be estimated by computing one-step TD errors, and decaying the weight of later TD errors at a rate of $\gamma\lambda\rho_t$. Implementing this online and incrementally, an \textit{eligibility trace} vector is maintained to track which state-action pairs led to the current step's TD error. The traces of earlier state-action pairs are decayed at each step by the afformentioned decay rate, and each action-value is adjusted by the current TD error weighted by the trace of the corresponding state-action pair.


Contrasting with $n$-step TD methods, the computational complexity of TD($\lambda$) control algorithms scales with the size of the environment, $|\mathcal{S}| \times |\mathcal{A}|$. That is, there is an environment-specific increase in complexity, but it no longer scales with the number of sampled rewards in the estimate of the return.

\section{PER-DECISION CONTROL VARIATES}
\label{sec:pdcv}


When trying to estimate the expectation of some variable $X$, control variates are often of the following form (Ross, 2013):
\begin{equation}
X^* = X + c(Y - \mathbb{E}[Y])
\label{eqn:cvform}
\end{equation}
where $Y$ is the outcome of another variable with a known expected value, and $c$ is a coefficient to be set. $X^*$ then has the following variance:
\begin{equation}
Var(X^*) = Var(X) + c^2Var(Y) + 2cCov(X,Y)
\label{eqn:cvvariance}
\end{equation}
From this, the variance can be minimized with the optimal coefficient $c^*$:
\begin{equation}
c^* = -\frac{Cov(X,Y)}{Var(Y)}
\label{eqn:cvcoeff}
\end{equation}
Suppose the $n$-step Sarsa algorithm samples the importance sampling-corrected $n$-step return, jointly samples the importance sampling-corrected action-value (through the sampled action), and computes the expected action-value under the target policy. We get the following estimate of this term of the multi-step return:
\begin{align}
\nonumber
(\rho_{t+1}&\hat{G}_{t+1:t+n})^* = \rho_{t+1}\hat{G}_{t+1:t+n} \\
&+ c\big(\rho_{t+1}Q(S_{t+1},A_{t+1}) - \mathbb{E}_{\pi}[Q(S_{t+1}, \cdot)]\big)
\label{eqn:returncvcoeff}
\end{align}
Under the assumption that the current estimates are accurate, the action-values represent the expected return. Due to this, the sampled reward sequence and the action-value are, in expectation, perfectly correlated. The covariance term in Equation \ref{eqn:cvcoeff} would then be the variance of the action-value due to the policy, and from this, a reasonable choice for the coefficient would be $-1$. This gives:
\begin{align}
\nonumber
(\rho_{t+1}&\hat{G}_{t+1:t+n})^* = \rho_{t+1}\hat{G}_{t+1:t+n} \\
&+ \mathbb{E}_{\pi}[Q(S_{t+1}, \cdot)] - \rho_{t+1}Q(S_{t+1},A_{t+1})
\label{eqn:returncv}
\end{align}
Substituting this estimate into the recursive definition of $n$-step Sarsa (\ref{eqn:recnsarsa}) and maintaining the same bootstrapping condition in Equation \ref{eqn:recnbootstrap} gives the following $n$-step return:
\begin{align}
\nonumber
\hat{G}_{t:t+n} &= R_{t+1} + \gamma \big(\rho_{t+1}\hat{G}_{t+1:t+n} \\
&+ \mathbb{E}_{\pi}[Q(S_{t+1}, \cdot)] - \rho_{t+1}Q(S_{t+1},A_{t+1})\big)
\label{eqn:nstepacv}
\end{align}
Because $\mathbb{E}_{\mu}[\mathbb{E}_{\pi}[Q(S_{t+1}, \cdot)] - \rho_{t+1}Q(S_{t+1},A_{t+1})] = 0$, the additional term does not introduce bias into the estimate. To provide an intuition of how it might reduce the variance in the estimate, we can consider some extreme cases of the importance sampling ratio. If $\rho_{t+1} = 0$, when the behaviour policy takes an action that the target policy would have never taken, it will bootstrap off of the expectation of its current estimates instead of cutting the return. If $\rho_{t+1}$ is much greater than $1$, an equivalent amount of its current action-value estimate is subtracted to compensate.

In the one-step case, the introduction of this control variate results in one-step Expected Sarsa's target:
\begin{align}
\nonumber
\hat{G}_{t:t+n} &= R_{t+1} + \gamma \big(\rho_{t+1}Q(S_{t+1},A_{t+1}) \\ \nonumber
&+ \mathbb{E}_{\pi}[Q(S_{t+1}, \cdot)] - \rho_{t+1}Q(S_{t+1},A_{t+1})\big) \\ \nonumber
\hat{G}_{t:t+n} &= R_{t+1} + \gamma \mathbb{E}_{\pi}[Q(S_{t+1}, \cdot)]
\end{align}
When applied at the bootstrapping step, it implicitly results in bootstrapping off of the expectation under the target policy as opposed to the importance sampling-corrected action-value. It can be viewed as an alternate generalization of Expected Sarsa to the multi-step setting, where the control variate is applied to the sampled reward sequence in addition to the bootstrapping step.

The control variate can be interpreted as performing an \textit{expectation correction} at each step based on current estimates. Each reward in the trajectory depends on the sampled action at each step, but the algorithm aims to learn the expectation across all possible trajectories under a policy. The importance sampling-corrected action-value is a closer estimate to the sampled return, as the agent knows which action resulted in the immediate reward at each step. Because of this, the action-value is like a guess of what the remainder of the sampled reward sequence will be, and the difference between that and the expectation across all actions provides a per-step estimate of the discrepancy between the sampled reward sequence and the expectation across all reward sequences from the current step onwards.

It can also be seen as implicitly performing adaptive $n$-step learning, adjusting the amount of information included based on how accurate its current estimates are. If we rearrange the $n$-step return:
\begin{align}
\nonumber
\hat{G}_{t:t+n} &= R_{t+1} + \gamma \mathbb{E}_{\pi}[Q(S_{t+1}, \cdot)] \\
&+ \gamma \big(\rho_{t+1}\hat{G}_{t+1:t+n}  - \rho_{t+1}Q(S_{t+1},A_{t+1})\big)
\label{eqn:adaptivenstep}
\end{align}
We get the one-step Expected Sarsa target, along with some difference between the actual sampled rewards and its current estimates. If the value estimates are poor, more rewards will be effectively included in the estimate, and vice-versa. If there is no stochasticity in the environment, it ends up approaching one-step Expected Sarsa as the estimates get close to the true value function.

If we follow similar steps in the state-value case, we arrive at the following $n$-step return with a per-decision control variate:
\begin{align}
\nonumber
\hat{G}_{t:t} &= V(S_t) \\ \nonumber
\hat{G}_{t:t+n} &= \rho_t(R_{t+1} + \gamma \hat{G}_{t+1:t+n}) + V(S_t) - \rho_t V(S_t) \\
\hat{G}_{t:t+n} &= \rho_t(R_{t+1} + \gamma \hat{G}_{t+1:t+n}) + (1 - \rho_t) V(S_t)
\label{eqn:nstepscv}
\end{align}
Of note, the state-value control variate disappears in the on-policy case, but the action-value one does not.

\section{RELATIONSHIP WITH EXISTING ALGORITHMS}
\label{sec:pdcvlambda}

If we substitute the $n$-step Sarsa return with the per-decision control variate (\ref{eqn:nstepacv}) into the definition of the $\lambda$-return in Equation \ref{eqn:lambdareturn}, we can rearrange it into a sum of one-step Expected Sarsa's TD errors:
\begin{align}
\nonumber
\hat{G}_{t} &= R_{t+1} + \gamma \mathbb{E}_{\pi}[Q(S_{t+1}, \cdot)] \\
\hat{G}_t^\lambda &= Q(S_t, A_t) + \sum_{k=t}^{\infty}{(\hat{G}_k - Q(S_k, A_k))\prod_{i=t+1}^{k}{\gamma \lambda \rho_i}}
\label{eqn:sumlexpsarsa}
\end{align}
This is equivalent to using the eligibility trace decay rate of Sarsa($\lambda$), but backing up the TD error of one-step Expected Sarsa. That is, in the space of action-value TD($\lambda$) algorithms, having the one-step estimates of the return bootstrap off of the expectation under the target policy implicitly induces this per-decision control variate in the corresponding $n$-step returns.

An existing algorithm that also uses one-step Expected Sarsa's TD error in its $\lambda$-return is the Tree-backup($\lambda$) algorithm (Precup et al., 2000). Denoting $\pi_t = \pi(S_t, A_t)$, Tree-backup($\lambda$) is characterized by the following equations:
\begin{align}
\nonumber
\hat{G}_{t} &= R_{t+1} + \gamma \mathbb{E}_{\pi}[Q(S_{t+1}, \cdot)] \\
\hat{G}_t^\lambda &= Q(S_t, A_t) + \sum_{k=t}^{\infty}{(\hat{G}_k - Q(S_k, A_k))\prod_{i=t+1}^{k}{\gamma \lambda \pi_t}}
\label{eqn:sumltb}
\end{align}
If we look at $n$-step Tree-backup's estimate of the return, we can show that it also includes the expectation correction terms: 
\begin{align}
\nonumber
\hat{G}_{t:t+n} &= R_{t+1} + \gamma (\pi_{t+1}\hat{G}_{t+1:t+n}  \\ \nonumber
&+ \sum_{a \neq A_{t+1}}{\pi(S_{t+1}, a) Q(S_{t+1}, a)} \\ \nonumber
\hat{G}_{t:t+n} &= R_{t+1} + \gamma (\pi_{t+1}\hat{G}_{t+1:t+n}  \\
&+\mathbb{E}_{\pi}[Q(S_{t+1}, \cdot)] - \pi_{t+1}Q(S_{t+1},A_{t+1}))
\label{eqn:nsteptb}
\end{align}
The estimate takes some portion of the sampled reward sequence, and the difference between the expectation under the target policy and an equivalent portion of the sampled action-value estimate.

The introduction of the control variates with the afformentioned choice of the control variate parameter results in an instance of a doubly robust estimator. The use of doubly-robust estimators in off-policy policy evaluation has been investigated by Jiang et al. (2016) and Thomas et al. (2016). However, results when applying the approaches in an online, model-free setting, as well as its view as a multi-step generalization of Expected Sarsa, appear to be novel.

Harutyunyan et al. (2016) has acknowledged the implicit introduction of these terms when using the expectation form of the TD error in action-value TD($\lambda$) algorithms. However, their work investigated the off-policy correcting effects of including the difference between the expectation under the target policy with an action-value sampled from the behaviour policy (without importance sampling corrections). This work focuses on the effect of explicitly including the additional terms, with importance sampling, in the $n$-step setting for both on and off-policy TD learning.


In the state-value setting, combining Equations \ref{eqn:nstepscv} and \ref{eqn:lambdareturn} gives the following $\lambda$-return:
\begin{align}
\nonumber
\hat{G}_{t} &= R_{t+1} + \gamma V(S_{t+1}) \\
\hat{G}_t^\lambda &= V(S_t) + \rho_t\sum_{k=t}^{\infty}{(\hat{G}_k - V(S_k))\prod_{i=t+1}^{k}{\gamma \lambda \rho_i}}
\label{eqn:sumltd}
\end{align}
which is an intuitive generalization of off-policy per-decision importance sampling for state-values, having an additional importance sampling correction term for the first reward in the sequence. It can be seen that the inclusion of an action-dependent trace decay rate scaling a TD error, as opposed to the return estimate alone, implicitly induces the state-value control variate in the $n$-step estimate of the return.

\section{EXPERIMENTS}
\label{sec:experiments}

In this section, we focus on the action-value setting and investigate the performance of $n$-step Sarsa with the per-decision control variate (denoted as \textit{$n$-step CV Sarsa}) on three problems. The first two are multi-step prediction tasks in a tabular environment, one being off-policy and one being on-policy. The remaining one is a control problem involving function approximation, evaluating the performance of $n$-step CV Sarsa beyond the tabular setting, as well as how it handles a changing (greedifying) policy.

\begin{figure}[h]
	\centering
	\includegraphics[width=0.5735549926\linewidth]{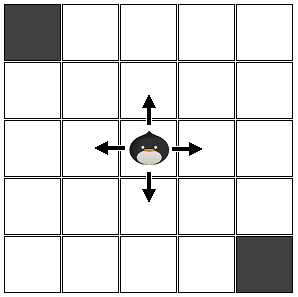}
	\caption{5$\times$5 Grid World environment. It was set up as an on and off-policy multi-step prediction task where the goal was to estimate the expected return under the target policy as accurately as possible.}
	\label{fig:5gwenv}
\end{figure}

\begin{figure*}[h]
	\includegraphics[width=0.5\linewidth]{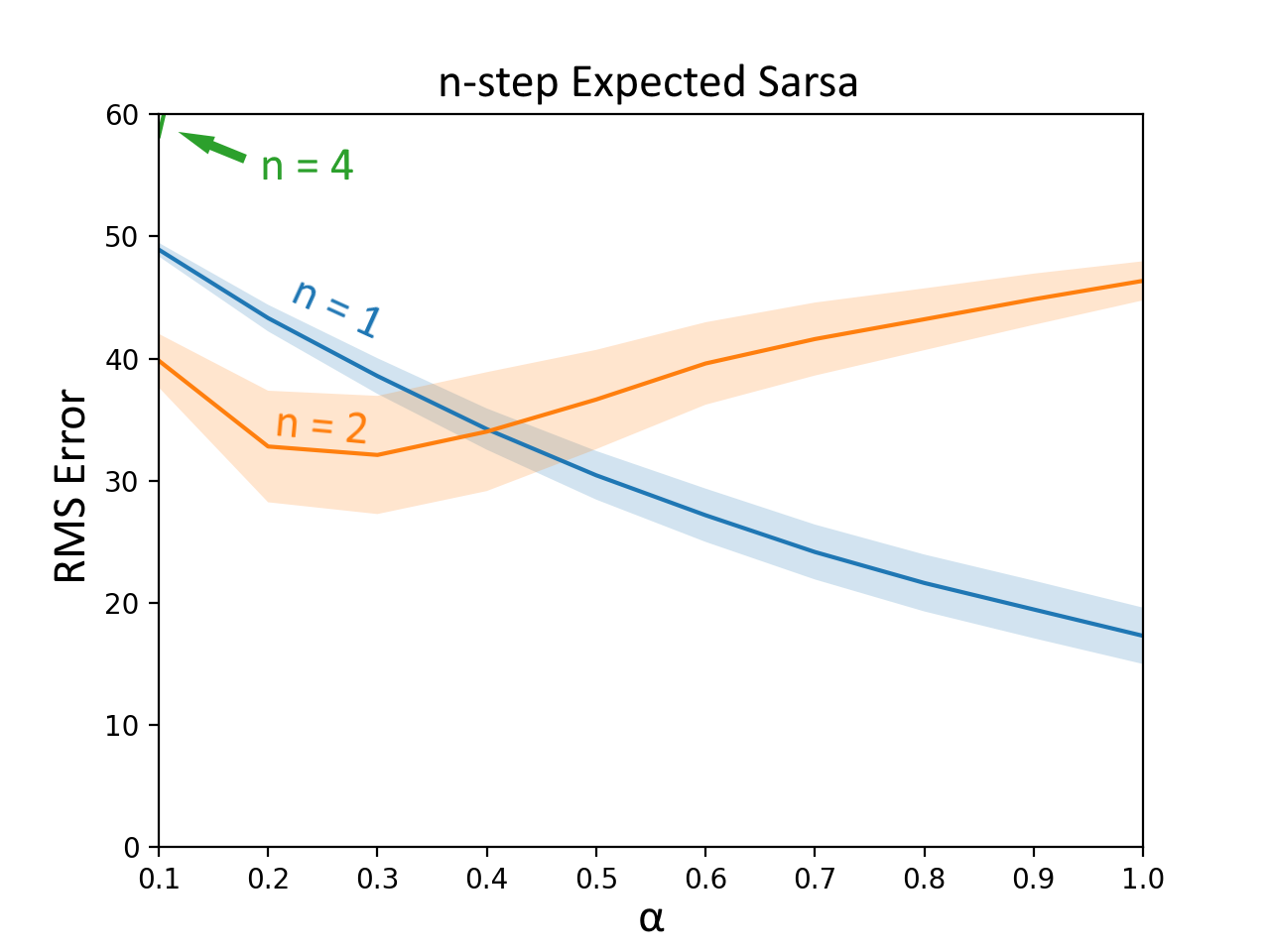}
	\includegraphics[width=0.5\linewidth]{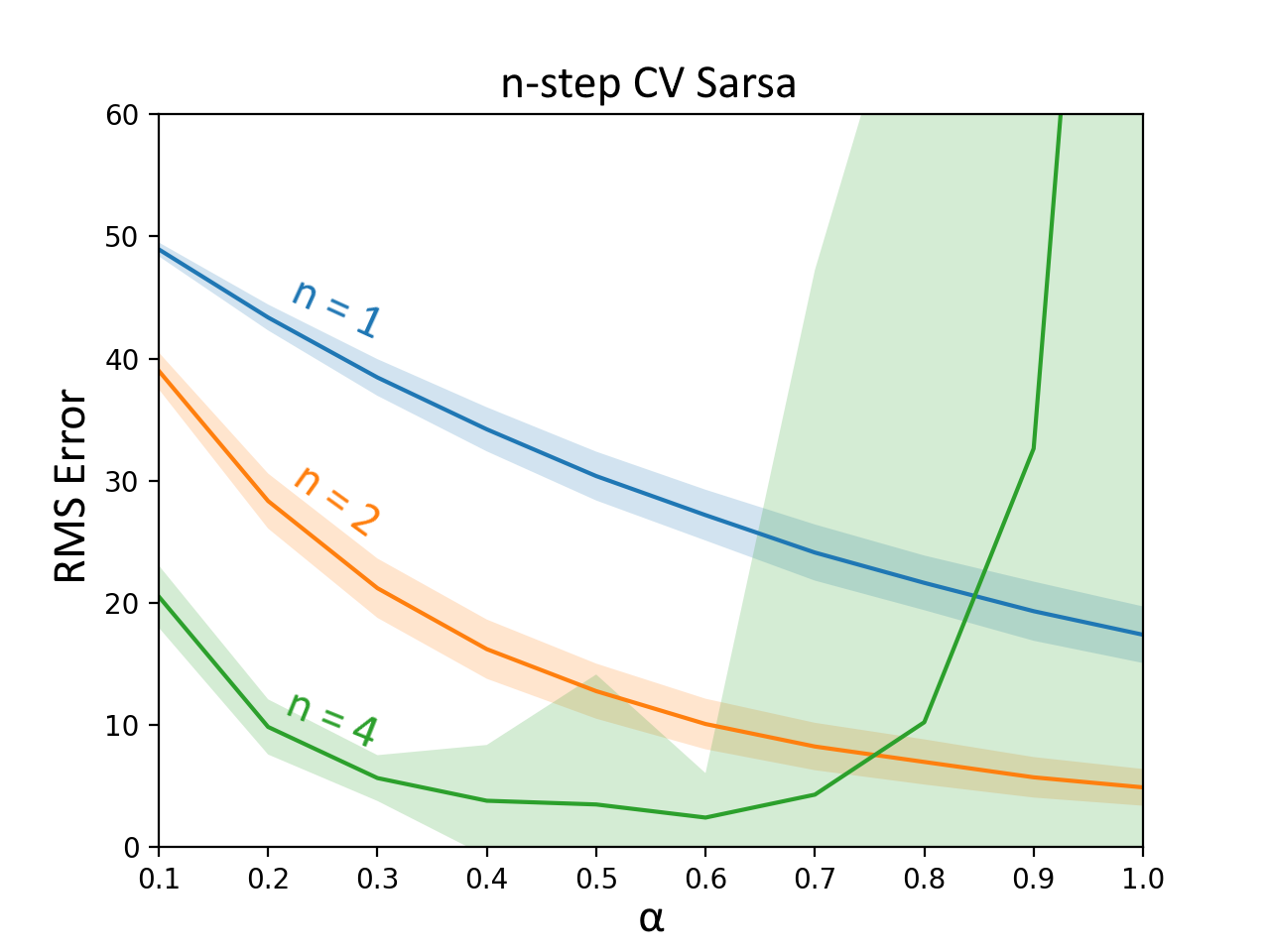}
	\caption{5x5 Grid World off-policy prediction results. The plot shows the performance of various parameter settings of each algorithm in terms of RMS error after 200 episodes in the learned value function. The shaded region corresponds to one standard deviation, and the results are averaged over 1000 runs.}
	\label{fig:5gw_cv_off}
\end{figure*}

Since $n$-step CV Sarsa ends up bootstrapping off of the expectation over action-values at the end of the reward sequence, we compare the algorithm to $n$-step Expected Sarsa as characterized by Equation \ref{eqn:nstepexpsarsa}. This allows for examining the effects of the control variate being applied to each reward in the reward sequence in an online and incremental setting.


\subsection{5$\times$5 GRID WORLD}
\label{sec:gridworld}
The \textit{5$\times$5 Grid World} is a 2-dimensional grid world having terminal states in two opposite corners. The actions consist of 4-directional movement, and moving into a wall transitions the agent to the same state. The agent starts in the center, and a reward of $-1$ is received at each transition. Experiments were run in both the off-policy and on-policy settings with no discounting ($\gamma = 1$), and the root-mean-square (RMS) error between the learned value function and the true value function were compared.

\subsubsection{Off-policy Prediction}
\label{sec:gridworldoff}

For the off-policy experiments, the target policy would move north with probability $1 - \epsilon$, and select a random action equiprobably otherwise. $\epsilon$ was set to 0.5, and the behaviour policy was equiprobable random for all states. A parameter study was done for 1, 2, and 4 steps, and the RMS error was measured after 200 episodes. The results are averaged over 1000 runs, and can be seen in Figure \ref{fig:5gw_cv_off}.

It can be seen that 2-step Expected Sarsa only outperforms 1-step Expected Sarsa for a very limited range of parameters, but is worse otherwise. 4-step Expected Sarsa was unable to learn for most parameter settings. When the control variate is applied to each reward, we can see that 2-step CV Sarsa outperforms 1-step Expected Sarsa for all parameter settings, and the variance is reduced relative to 2-step Expected Sarsa. Furthermore, 4-step CV Sarsa ends up being able to learn, and can outperform 2-step CV Sarsa for a reasonably wide range of parameters.

\subsubsection{On-policy Prediction}
\label{sec:gridworldon}

\begin{figure*}[h]
	\includegraphics[width=0.5\linewidth]{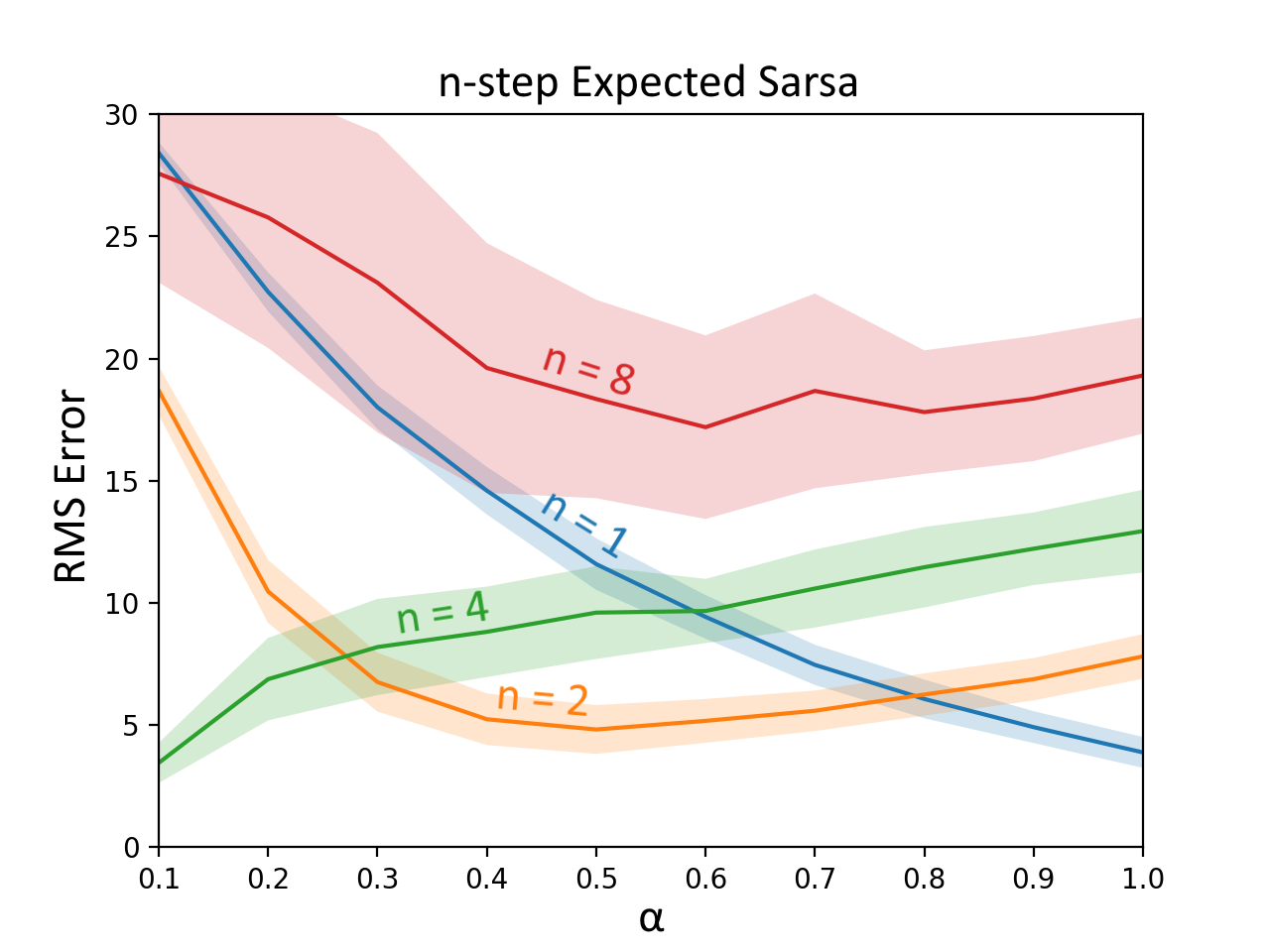}
	\includegraphics[width=0.5\linewidth]{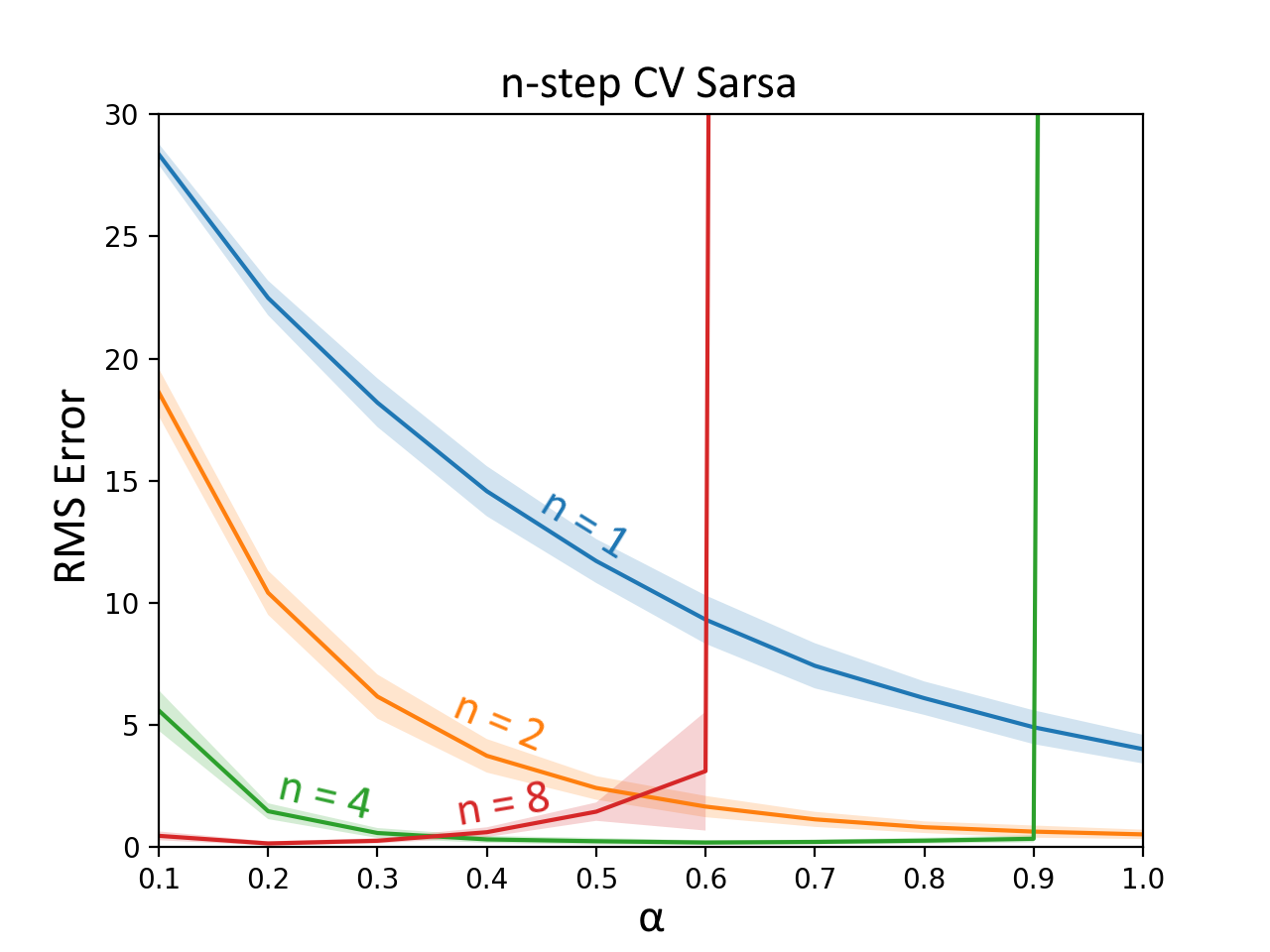}
	\caption{5x5 Grid World on-policy prediction results. The plot shows the performance of various parameter settings of each algorithm in terms of RMS error after 200 episodes in the learned value function. The shaded region corresponds to one standard deviation, and the results are averaged over 1000 runs.}
	\label{fig:5gw_cv_on}
\end{figure*}

In the on-policy case, the target policy and behaviour policy were both equiprobable random for all states. The parameters tested are identical to the off-policy experiment with the addition of 8-step instances of each algorithm. The RMS error was measured after 200 episodes, and are also averaged over 1000 runs. The results are summarized in Figure \ref{fig:5gw_cv_on}.

2-step Expected Sarsa ends up performing better than 1-step Expected Sarsa for a wider range of parameters than in the off-policy case, but the best parameter settings for each perform similarly. Further increasing the number of steps results in relatively poor performance, and doesn't do better than the best parameter setting of 1-step Expected Sarsa. Looking at $n$-step CV Sarsa, we can see that performance is drastically improved for all tested settings of $n$. Of note, while introducing the per-decision control variate resulted in lower variance for a reasonable range of parameters, assumptions were made regarding the accuracy of the value function when setting the control variate parameter $c$ in Equation \ref{eqn:returncvcoeff}. If the number of steps $n$ and the step size $\alpha$ get too large, it can result in larger variance and divergence on parameter settings where $n$-step Expected Sarsa did not diverge. We did not investigate alternate methods of setting the control variate parameter in this work.

\subsection{MOUNTAIN CAR}
\label{sec:mountaincar}

To show that this use of control variates is compatible with function approximation, we ran experiments on \textit{mountain car} as described by Sutton and Barto (1998). A reward of $-1$ is received at each step, and there is no discounting.

\begin{figure}[h]
	\centering
	\includegraphics[width=1.0\linewidth]{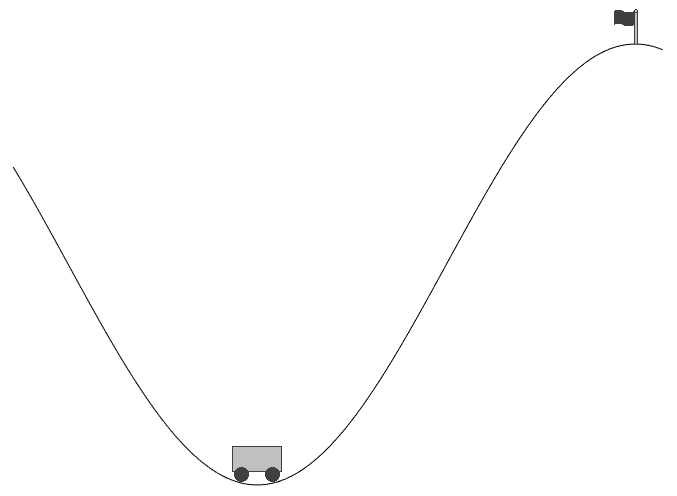}
	\caption{The mountain car environment (Sutton \& Barto, 1998). The agent starts at a random location in the valley, receives a reward of $-1$ at each step, and its goal is to drive past the flag in as few steps as possible.}
	\label{fig:mcarenv}
\end{figure}

Because the environment's state space is continuous, we used \textit{tile coding} (Sutton \& Barto, 1998) to produce a feature representation for use with linear function approximation. The tile coder used 16 tilings, an asymmetric offset by consecutive odd numbers, and each tile covered $1/8$-th of the feature space in each direction.

\begin{figure}[h]
	\includegraphics[width=\linewidth]{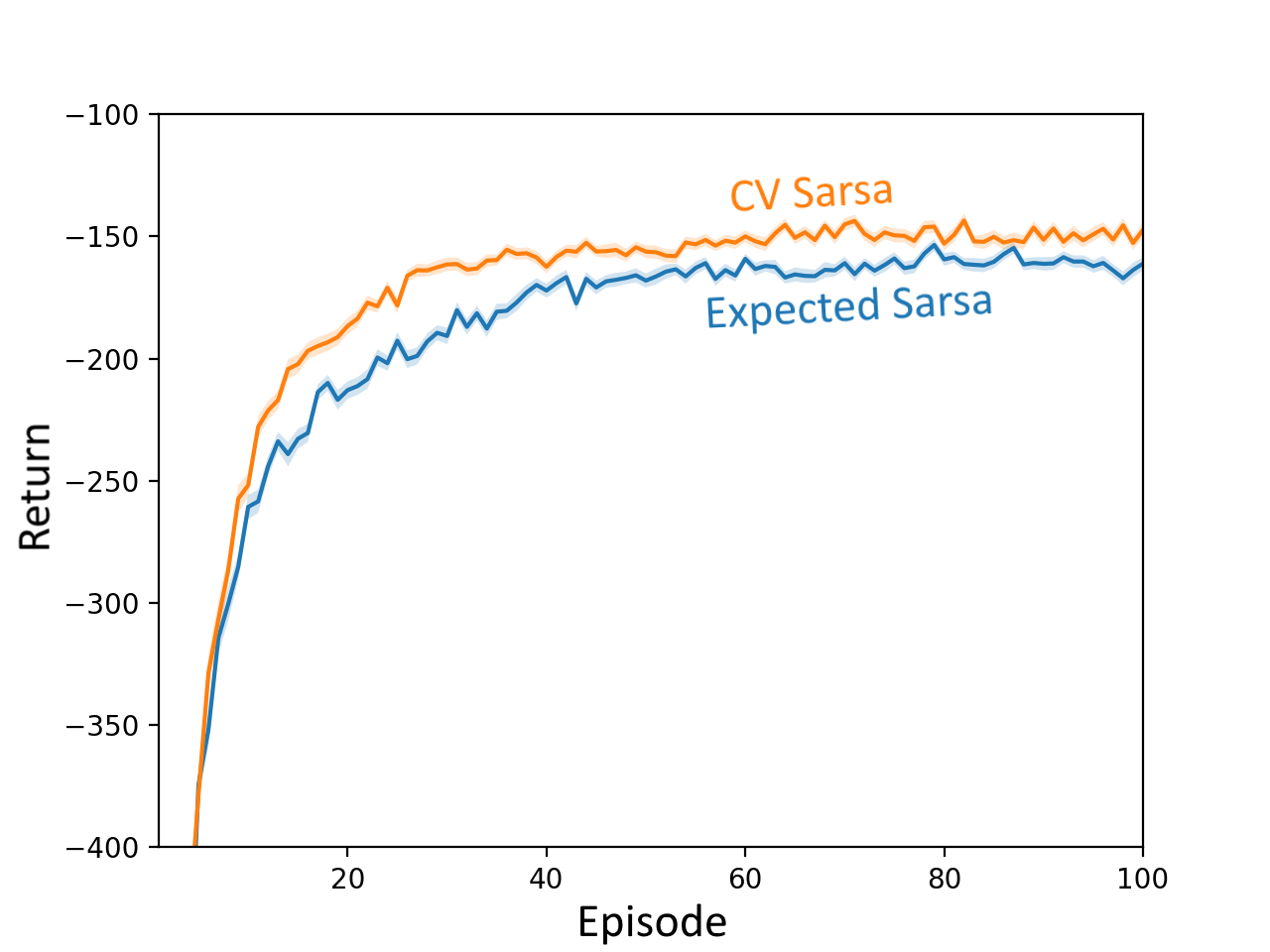}
	\caption{Mountain Car on-policy control results. The plot shows the return per episode of the best parameter setting of each algorithm in terms of the mean return over all episodes. The shaded region corresponds to one standard error, and the results are averaged over 100 runs.}
	\label{fig:mc_on}
\end{figure}

We compared $n$-step Expected Sarsa and $n$-step CV Sarsa with 1, 2, 4, and 8 steps across different step sizes $\alpha$. Each algorithm learned on-policy with an $\epsilon$-greedy policy which selects an action greedily with respect to its value function with probability $1 - \epsilon$, and selected a random action equiprobably otherwise. In this experiment, $\epsilon$ was set to 0.1. We measured the return per episode up to 100 episodes, and averaged the results over 100 runs. The results for the best parameter setting for each algorithm can be found in Figure \ref{fig:mc_on}.

The two algorithms showed a similar trend in the parameters as in the 5$\times$5 Grid World environment, but were less pronounced. This is likely due to not requiring accurate value function estimates to perform the task well, and the control variate having less of an effect with greedier target policies, because  $\mathbb{E}_{\pi}[Q(S_t,\cdot)]$ gets relatively close to $Q(S_t, A_t)$. Despite this, as seen in the results for the best parameter settings, $n$-step CV Sarsa still outperforms $n$-step Expected Sarsa on this task.

\section{DISCUSSION}
\label{sec:discussion}

From our experiments, $n$-step CV Sarsa appears to be an improved multi-step generalization of Expected Sarsa. In both on and off-policy prediction tasks on the 5$\times$5 Grid World environment, it generally resulted in lower variance as well as considerably lower error in the estimates compared to $n$-step Expected Sarsa, an algorithm which can be interpreted as only applying the control variate at the bootstrapping step.  Moreover, when used on a continuous state space control problem with function approximation, applying the control variate on a per-reward level still resulted in greater performance in terms of average return per episode.

Despite the improvement on most of the tested parameter settings, the results also showed that the addition of the per-decision control variates can cause learning to diverge for large $n$ and large step size $\alpha$, even on settings where $n$-step Expected Sarsa did not diverge. It is suspected that this is due to assuming the estimates are accurate when setting the control variate parameter in Equation \ref{eqn:returncvcoeff}. This was not further investigated, but it could be an avenue for future work.

While our results focused on the action-value per-decision control variate, other experiments not included in this paper showed that the state-value per-decision control variate in Equation \ref{eqn:nstepscv} can also be applied in the off-policy action-value setting. It resulted in performance in between that of $n$-step Expected Sarsa and $n$-step CV Sarsa, supporting that it is beneficial to add it, but better to use the action-value control variate if the agent is learning action-values.

\section{CONCLUSIONS}
\label{sec:conclusions}

In this paper, we presented a way to derive per-decision control variates in both state-value and action-value $n$-step TD methods. The state-value control variate is only present in the off-policy setting, but the action-value control variate affects both on and off-policy learning. In the action-value case, applying the per-decision control variate results in an alternative multi-step extension of Expected Sarsa. With this control variate perspective, the existing $n$-step Expected Sarsa algorithm can be interpreted as only applying a control variate at the bootstrapping step, when it can be applied to the sampled reward sequence as well. Our results on prediction and control problems show that applying it on a per-decision level can greatly improve the accuracy of the learned value function, and consequently perform better when doing TD control.

We also showed how the per-decision control variates relate to TD($\lambda$) algorithms. This provided insight on how minor adjustments in the TD($\lambda$) space can implicitly induce these per-decision control variates in the underlying $n$-step returns, resulting in a more unified view of per-decision multi-step TD methods.

Our experiments were limited to the $n$-step TD setting without eligibility traces, and focused on learning action-values. We only considered a naive setting of the control variate scaling parameter $c$, when our results suggest that the way we set it can negatively affect learning for a few (relatively extreme) parameter combinations. Perhaps insight from the analytical optimal coefficient in Equation \ref{eqn:cvcoeff} can be used to adapt the control variate online to further improve performance.

\newpage

\subsubsection*{Acknowledgements}

The authors thank Yi Wan for insights and discussions contributing to the results presented in this paper, and the entire Reinforcement Learning and Artificial Intelligence research group for providing the environment to nurture and support this research. We gratefully acknowledge funding from Alberta Innovates -- Technology Futures, Google Deepmind, and from the Natural Sciences and Engineering Research Council of Canada.

\subsubsection*{References}

\hangindent=1em \hangafter=1
Harutyunyan, A., Bellemare, M. G., Stepleton, T., and Munos, R. (2016). Q($\lambda$) with off-policy corrections. \textit{arXiv:1509.05172}.

\hangindent=1em \hangafter=1
Jaakola, T., Jordan, M. I., and Singh, S. P. (1994). On the convergence of stochastic iterative dynamic programming algorithms. \textit{Neural Computation 6}(6), 1185-1201.

\hangindent=1em \hangafter=1
Jiang, N., and Li, L. (2016). Doubly Robust Off-policy Value Evaluation for Reinforcement Learning. \textit{Proceedings of The 33rd International Conference on Machine Learning}, in PMLR 48:652-661.

\hangindent=1em \hangafter=1
Mahmood, A. R., Yu, H., and Sutton, R. S. (2017). Multi-step off-policy learning without importance sampling ratios. \textit{arXiv:1702.03006}.

\hangindent=1em \hangafter=1
Munos, R., Stepleton, T., Haruytunyan, A., and Bellemare, M. G. (2016). Safe and efficient off-policy reinforcement learning. \textit{arXiV:1606.02647}.

\hangindent=1em \hangafter=1
Precup, D., Sutton, R. S., and Singh, S. (2000). Eligibility traces for off-policy policy evaluation. In \textit{Proceedings of the 17th International Conference on Machine Learning}, pp. 759-766. Morgan Kaufmann.

\hangindent=1em \hangafter=1
Ross, S. M. (2013). \textit{Simulation}. San Diego: Academic Press.

\hangindent=1em \hangafter=1
Rummery, G. A. (1995). \textit{Problem Solving with Reinforcement Learning}. PhD Thesis, Cambridge University.

\hangindent=1em \hangafter=1
Rummery, G. A., and Niranjan, M. (1994). On-line Q-learning using connectionist systems. Technical Report CUEF/F-INFENG/TR 166, Engineering Department, Cambridge University.

\hangindent=1em \hangafter=1
Sutton, R. S. (1988). Learning to predict by the methods of temporal differences. \textit{Machine Learning 3}, 9-44.

\hangindent=1em \hangafter=1
Sutton, R. S. (1996). Generalization in reinforcement learning: Successful examples using sparse coarse coding. In Touretzky, D. S. and Hasselmo, M. E. (eds.), \textit{Advances in Neural Information Processing Systems 8}, pp. 1038-1044. MIT Press.

\hangindent=1em \hangafter=1
Sutton, R. S., and Barto, A. G. (1998). \textit{Reinforcement Learning: An Introduction}. MIT Press, Cambridge, Massachusetts.

\hangindent=1em \hangafter=1
Sutton, R. S., and Barto, A. G. (2018). \textit{Reinforcement Learning: An Introduction} (2nd ed.). Manuscript in preparation.

\hangindent=1em \hangafter=1
Thomas, P. and Brunskill, E. (2016). Data-Efficient Off-Policy Policy Evaluation for Reinforcement Learning. \textit{Proceedings of The 33rd International Conference on Machine Learning}, in PMLR 48:2139-2148.

\hangindent=1em \hangafter=1
van Seijen, H., van Hasselt, H., Whiteson, S., and Wiering, M. (2009). A theoretical and empirical analysis of expected sarsa. In \textit{Proceedings of the IEEE Symposium on Adaptive Dynamic Programming and Reinforcement Learning}, pp. 177-184.

\hangindent=1em \hangafter=1
Watkins, C. J. C. H. (1989). \textit{Learning from Delayed Rewards}. PhD thesis, Cambridge University.

\end{document}